\pdfoutput=1

\documentclass[twocolumn]{bmcart}

\usepackage{amsthm,amsmath} 
\usepackage[caption=false]{subfig}
\usepackage{dblfloatfix}
\usepackage{graphicx}
\usepackage[utf8]{inputenc} 
\usepackage{url}
\usepackage{booktabs}
\usepackage{xcolor}
\usepackage{xspace}
\usepackage[normalem]{ulem}

\makeatletter
\DeclareRobustCommand\onedot{\futurelet\@let@token\@onedot}
\def\@onedot{\ifx\@let@token.\else.\null\fi\xspace}

\def\eg{\emph{e.g}\onedot} 
\def\ie{\emph{i.e}\onedot} 
 
\def\etc{\emph{etc}\onedot} 
 
\def\etal{\emph{et al}\onedot}
\makeatother

\def\para#1{\smallskip\noindent{\bf{#1}}}

\startlocaldefs
\endlocaldefs

\begin{document}

\begin{frontmatter}

\begin{fmbox}
\dochead{}


\title{Structure-from-Motion using Dense CNN Features with Keypoint Relocalization}


\author[
   addressref={aff1},                   
   email={widya.a.aa@m.titech.ac.jp}   
]{\inits{AR}\fnm{Aji Resindra} \snm{Widya}}
\author[
   addressref={aff1},
   email={torii@ctrl.titech.ac.jp}
]{\inits{AT}\fnm{Akihiko} \snm{Torii}}
\author[
   addressref={aff1},
   email={mxo@sc.e.titech.ac.jp }
]{\inits{MXO}\fnm{Masatoshi} \snm{Okutomi}}


\address[id=aff1]{
  \orgname{Tokyo Institute of Technology}, 
  \street{O-okayama, Meguro-ku},                     %
  \postcode{152-8550}                                
  \city{Tokyo},                              
  \cny{Japan}                                    
}



\begin{artnotes}
\end{artnotes}


\begin{abstractbox}

\begin{abstract} 
Structure from Motion (SfM) using imagery that involves extreme appearance changes is yet a challenging task due to a loss of feature repeatability. Using feature correspondences obtained by matching densely extracted convolutional neural network (CNN) features significantly improves the SfM reconstruction capability. However, the reconstruction accuracy is limited by the spatial resolution of the extracted CNN features which is not even pixel-level accuracy in the existing approach. Providing dense feature matches with precise keypoint positions is not trivial because of memory limitation and computational burden of dense features. To achieve accurate SfM reconstruction with highly repeatable dense features, we propose an SfM pipeline that uses dense CNN features with relocalization of keypoint position that can efficiently and accurately provide pixel-level feature correspondences. Then, we demonstrate on the Aachen Day-Night dataset that the proposed SfM using dense CNN features with the keypoint relocalization outperforms a state-of-the-art SfM (COLMAP using RootSIFT) by a large margin. 
\end{abstract}




\end{abstractbox}
\end{fmbox}

\end{frontmatter}



\section{Introduction}
Structure from Motion (SfM) is getting ready for 3D reconstruction only using images, thanks to off-the-shelf softwares~\cite{pix,agisoft,photomodeler} and open-source libraries~\cite{fuhrmann_mve-multi-view_2014,sweeney_theia:_2015,schonberger_structure--motion_2016,schonberger_pixelwise_2016,wilson2014robust,moulon_openmvg:_2016,cui2017hsfm}. 
They provide impressive 3D models, especially, when targets are captured from many viewpoints with large overlaps. The state-of-the-art SfM pipelines, in general, start with extracting local features~\cite{lowe_distinctive_2004,mikolajczyk2004scale,kadir2004affine,tuytelaars2004matching,arandjelovic2012three,dong2015domain,yi_lift:_2016} and matching them across images, followed by pose estimation, triangulation, and bundle adjustment~\cite{snavely_modeling_2008,agarwal_reconstructing_2010,agarwal_building_2011}. The performance of local features and their matching, therefore, is crucial for 3D reconstruction by SfM. 

In this decade, the performance of local features, namely, SIFT~\cite{lowe_distinctive_2004} and its variants~\cite{dong2015domain,morel2009asift,ke2004pca,abdel2006csift,bay2006surf} are validated on 3D reconstruction as well as many other tasks~\cite{csurka2004visual,lazebnik2006beyond,chong2009simultaneous}.
The local features give promising matches for well-textured surfaces/objects but significantly drop its performance for matching weakly-textured objects~\cite{hinterstoisser2012gradient}, repeated patterns~\cite{torii2013visual}, extreme changes of viewpoints~\cite{morel2009asift,mishkin2015mods,taira2016robust} and illumination change~\cite{torii_24/7_2015,radenovic_dusk_2016} because of degradation in repeatability of feature point (keypoint) extraction~\cite{morel2009asift,taira2016robust}.
This problem can be mitigated by using densely detected features on a regular grid~\cite{bosch_image_2007,liu2016sift} but their merit is only demonstrated in image retrieval~\cite{torii_24/7_2015,zhao2013oriented} or image classification tasks~\cite{lazebnik2006beyond,bosch_image_2007} that use the features for global image representation and do not require one-to-one feature correspondences as in SfM. 

Only recently, SfM with densely detected features are presented in~\cite{sattler2017benchmarking}. 
DenseSfM~\cite{sattler2017benchmarking} uses convolutional neural network (CNN) features as densely detected features, \ie, it extracts convolutional layers of deep neural network~\cite{simonyan2014very} and converts them as feature descriptors of keypoints on a grid pattern (Section~\ref{sec:denseextract}).
As the main focus of \cite{sattler2017benchmarking} is camera localization, the SfM architecture including neither dense CNN feature description and matching nor its 3D reconstruction performance is not studied in detail.

\para{Contribution.} In this work, we first review the details of the SfM pipeline with dense CNN feature extraction and matching. We then propose a keypoint relocalization that uses the structure of convolutional layers (Section~\ref{sec:matching}) to overcome keypoint inaccuracy on the grid resolution and computational burden of dense feature matching. Finally, the performance of SfM with dense CNN feature using the proposed keypoint relocalization is evaluated on Aachen Day-Night~\cite{sattler2017benchmarking} dataset and additionally on Strecha~\cite{strecha_benchmarking_2008} dataset. 
  
 \begin{figure*}[tb]
\centering
\includegraphics[width=0.96\textwidth]{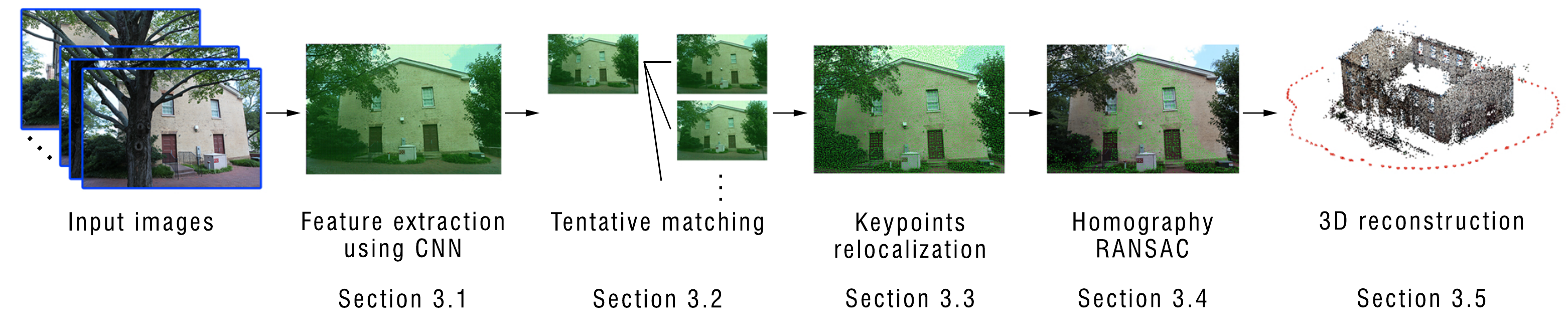}
\caption{{\bf Pipeline of the proposed SfM using dense CNN features with keypoint relocalization}. Our SfM starts from dense feature extraction (Section~\ref{sec:denseextract}), feature matching (Section~\ref{sec:matching}), the proposed keypoint relocalization (Section~\ref{sec:relocalization}), feature verification using RANSAC with multiple homographies (Section~\ref{sec:ransac}) followed by 3D reconstruction (Section~\ref{sec:reconstruction}).}
\label{fig:flowchart}
\end{figure*}

\section{Related work}
\label{ref:related}
\para{SfM and VisualSLAM.} 
The state-of-the-art SfM is divided into a few mainstream pipelines: incremental (or sequential)~\cite{fuhrmann_mve-multi-view_2014,schonberger_structure--motion_2016,wu_towards_2013}, global~\cite{wilson2014robust,moulon_openmvg:_2016,cui2015global}, and hybrid~\cite{cui2017hsfm,magerand2017practical}.

In VisualSLAM approaches, namely, LSD-SLAM~\cite{engel2014lsd} and DTAM~\cite{newcombe2011dtam} repeat camera pose estimation based on selected keyframe and (semi-)dense reconstruction using the pixel-level correspondences in real-time.
These methods are particularly designed to work with video streams, \ie, short baseline camera motion, but not with general wide-baseline camera motion.

Recently,~\cite{sattler2017benchmarking} introduces CNN-based DenseSfM that adopts densely detected and described features. But, their SfM uses fixed poses and intrinsic parameters of reference images in evaluating the performance of query images localization. They also do not address keypoint inacuraccy of CNN features. Therefore, it remains as an open challenge.

\para{Feature points.}
The defacto standard local feature, SIFT~\cite{lowe_distinctive_2004}, is capable of matching images under viewpoint and illumination changes thanks to scale and rotation invariant keypoint patches described by histograms of the oriented gradient. 
ASIFT~\cite{morel2009asift} and its variants~\cite{mishkin2015mods,taira2016robust} explicitly generate synthesized views in order to improve repeatability of keypoint detection and description under extreme viewpoint changes. 

An alternative approach to improve feature matching between images across extreme appearance changes is to use densely sampled features from images. 
Densely detected features are often used in multi-view stereo~\cite{furukawa2015multi} with DAISY~\cite{tola_daisy:_2010}, or image retrieval and classification~\cite{liu2016sift,tuytelaars2010dense} with Dense SIFT~\cite{bosch_image_2007}. 
However, dense features are not spotlighted in the task of one-to-one feature correspondence search under unknown camera poses due to its loss of scale, rotation invariant, inaccuracy of localized keypoints, and computational burden.

\para{CNN features.}
Fischer~\etal~\cite{fischer_descriptor_2014} reported that, given feature positions, descriptors extracted from CNN layer have better matchability compared to SIFT~\cite{lowe_distinctive_2004}. More recently, Schonberger~\etal~\cite{schonberger_comparative_2017} also showed that CNN-based learned local features such as LIFT~\cite{yi_lift:_2016}, \textit{Deep-Desc}~\cite{simo2015discriminative}, and \textit{ConvOpt}~\cite{simonyan2014learning} have higher recall compared to SIFT~\cite{lowe_distinctive_2004} but still cannot outperform its variants, e.g., DSP-SIFT~\cite{dong2015domain} and SIFT-PCA~\cite{bursuc2015kernel}.

Those studies motivate us to adopt CNN architecture for extracting features from images and matching them for SfM as it efficiently outputs multi-resolution features and has potential to be improved by better training or architecture.

\section{The pipeline: SfM using dense CNN features with keypoint relocalization}
\label{sec:proposedmethod}
Our SfM using densely detected features mimics the state-of-the-art incremental SfM pipeline that consists of 
feature extraction (Section~\ref{sec:denseextract}), feature matching (Section~\ref{sec:matching} to~\ref{sec:ransac}), and incremental reconstruction (Section~\ref{sec:reconstruction}). 
Figure~\ref{fig:flowchart} overviews the pipeline. 
In this section, we describe each component while stating the difference to the sparse keypoint based approaches. 

\subsection{Dense feature extraction} 
\label{sec:denseextract}
Firstly, our method densely extracts the feature descriptors and their locations from the input image. In the same spirit of~\cite{arandjelovic_netvlad:_2016,Radenovic-ECCV16}, we input images in a modern CNN architecture~\cite{simonyan2014very,szegedy2015going,he2016deep} and use the convolutional layers as densely detected keypoints on a regular grid, \ie, cropping out the fully connected and softmax layers. In the following, we chose VGG-16~\cite{simonyan2014very} as the base network architecture and focus on the description tailored to it, but this can be replaced with other networks with marginal modification. 

\begin{figure*}[tb]
\includegraphics[width=0.9\textwidth]{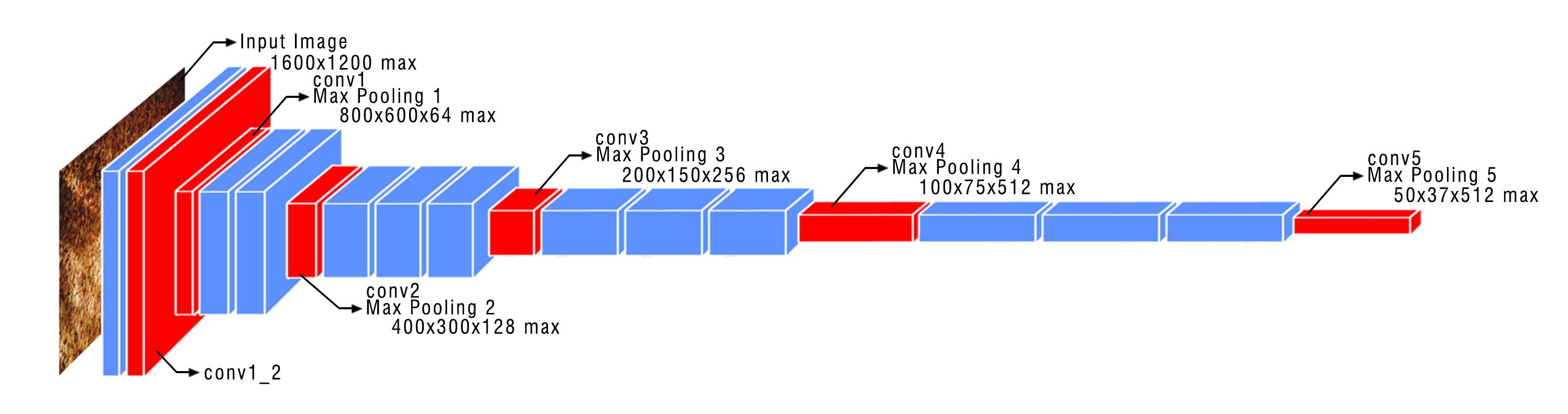}
\caption{{\bf Features extracted using CNN}. The figure summarizes blocks of convolutional layers of VGG-16 as an example of CNN architecture. Our SfM uses the layers colored in red as features. For example, given an input image of $1600 \times 1200$ pixels, we extract $256$ dimensional features of $200 \times 150$ spatial resolution from the conv3 max-pooling.}
\label{fig:VGG}
\end{figure*}

As illustrated in Figure~\ref{fig:VGG}, VGG-16~\cite{simonyan2014very} is composed of five max-pooling layers and 16 weight layers. 
We extract the max-pooling layers as dense features. As can be seen in Figure~\ref{fig:VGG}, the conv1 max-pooling layer is not yet the same resolution as the input image. We, therefore, also extract conv1\_2, one layer before the conv1 max-pooling layer, that has pixel-level accuracy.

\subsection{Tentative matching}
\label{sec:matching}
Given multi-level feature point locations and descriptors, tentative matching uses upper max-pooling layer (lower spatial resolution) to establish initial correspondences. This is motivated by that the upper max-pooling layer has a larger receptive field and encodes more semantic information~\cite{fischer_descriptor_2014,berkes2006analysis,zeiler2014visualizing} which potentially gives high matchability across appearance changes.
Having the lower spatial resolution is also advantageous in the sense of computational efficiency. 

For a pair of images, CNN descriptors are tentatively matched by searching their nearest neighbors (L2 distances) and refined by taking mutually nearest neighbors.
Note that the standard ratio test~\cite{lowe_distinctive_2004} removes too many feature matches as neighborhood features on a regularly sampled grid tend to be similar to each other. 

We perform feature descriptor matching for all the pairs of images or shortlisted images by image retrieval, \eg, NetVLAD~\cite{arandjelovic_netvlad:_2016}. 

\subsection{Keypoint relocalization}
\label{sec:relocalization}

The tentative matching using the upper max-pooling layers, \eg, conv5, generates distinctive correspondences but the accuracy of keypoint position is limited by their spatial resolution. This inaccuracy of keypoints can be mitigated by a coarse-to-fine matching from the extracted max-pooling layer up to conv1\_2 layer utilizing extracted intermediate max-pooling layers between them. 

For example, the matched keypoints found on the conv3 layer are transferred to the conv2 (higher spatial resolution) and new correspondences are searched only in the area constrained by the transferred keypoints. This can be repeated until reaching conv1\_2 layer. However, this naive coarse-to-fine matching generates too many keypoints that may lead to a problem in computational and memory usage in incremental SfM step, especially, bundle adjustment.

To generate dense feature matches with pixel-level accuracy while preserving their quantity, we propose a method of keypoint relocalization as follows. 

\begin{figure*}[tb]
\subfloat[]{
   \includegraphics[width=\linewidth]{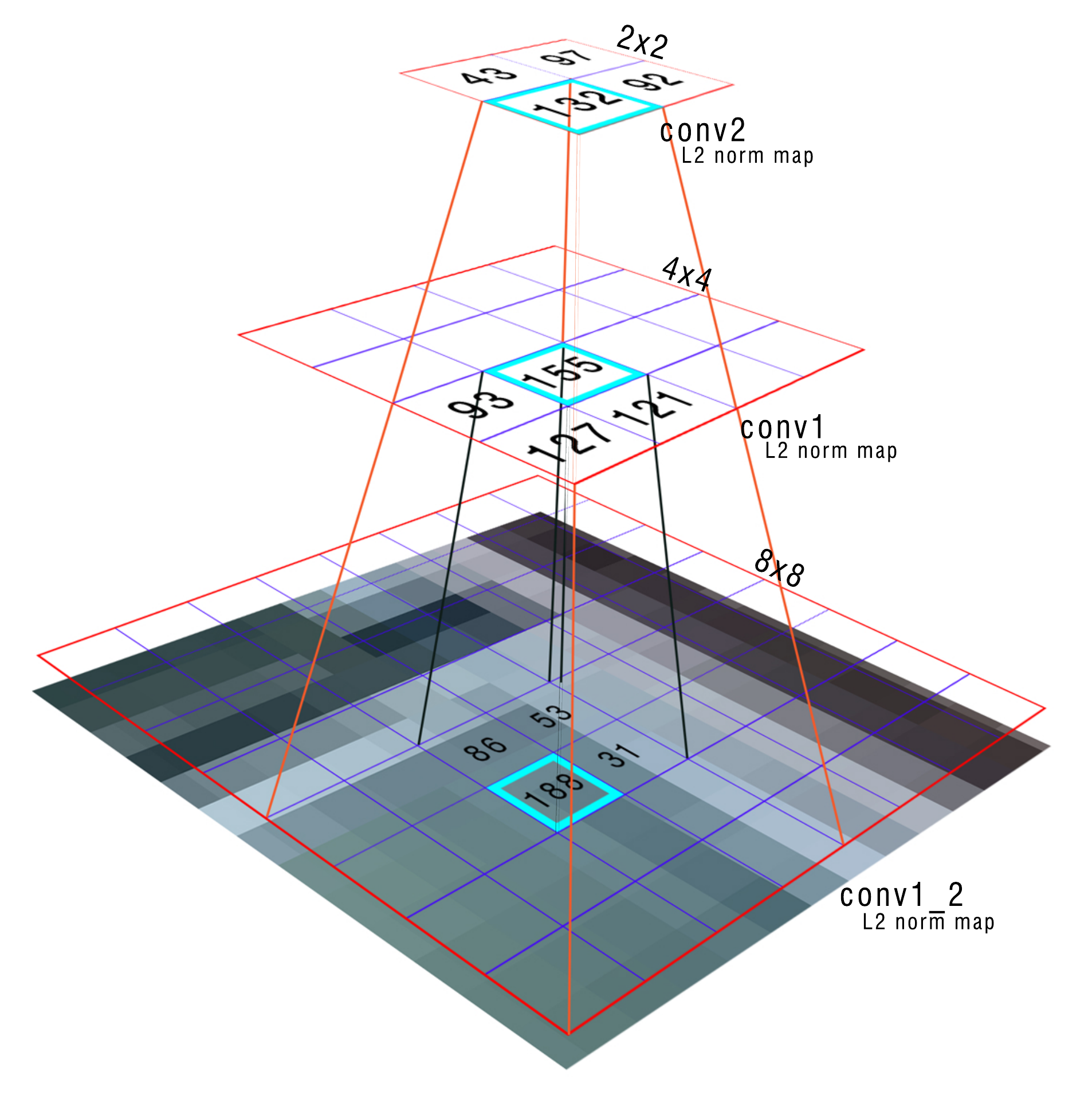}
 }
\subfloat[]{
   \includegraphics[width=0.8\linewidth]{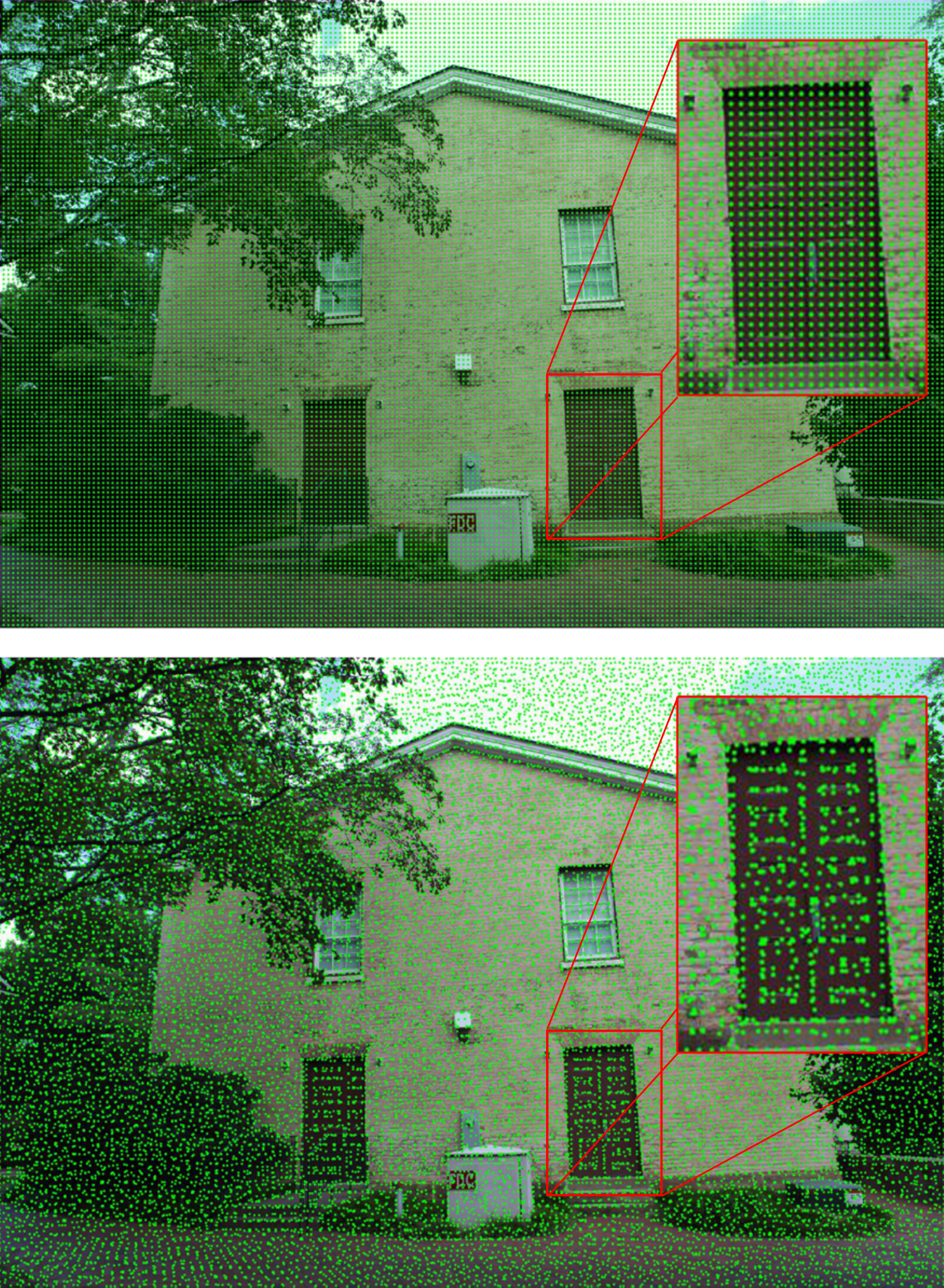}
 }
\caption{{\bf Keypoint relocalization}. 
(a) A keypoint on a sparser level is relocalized using a map computed from descriptors' L2 norm on an lower level which has higher spatial resolution. 
It is reassigned at the position on the lower level which has the largest value in the corresponding $K \times K$ neighborhood. By repeating this, the relocalized keypoint position in conv1\_2 has the accuracy as in the input image pixels. (b) The green dots show the extracted conv3 features points (top) and the result of our keypoint relocalization (bottom).}
\label{fig:reloc}

\end{figure*}

For each feature point at the current layer, we retrieve the descriptors on the lower layer (higher spatial resolution) in the corresponding $K \times K$ pixels\footnote{We use $K=2$ throughout the experiments.}. The feature point is relocalized at the pixel position that has the largest descriptor norm (L2 norm) in the $K \times K$ pixels. This relocalization is repeated until it reaches the conv1\_2 layer which has the same resolution as the input image (see also Figure~\ref{fig:reloc}).

\subsection{Feature verification using RANSAC with multiple homographies}
\label{sec:ransac}
Using all the relocated feature points, we next remove outliers from a set of tentative matches by Homography-RANSAC. 
We rather use a vanilla RAN\-SAC instead of the state-of-the-art spatial verification~\cite{Philbin07} by taking into account the spatial density of feature correspondences. 
To detect inlier matches lying on several planes, Homography-RANSAC is repeated while excluding the inlier matches of the best hypothesis. The RANSAC inlier/outlier threshold is set to be loose to allow features off the planes. 

\subsection{3D reconstruction}
\label{sec:reconstruction} 
Having all the relocalized keypoints filtered by RAN\-SAC, we can export them to any available pipelines that perform pose estimation, point triangulation, and bundle adjustment. 

Dense matching may produce many confusing feature matches on the scene with many repetitive structures, \eg, windows, doors, pillars, \etc. In such cases, we keep only the $N$ best matching image pairs for each image in the dataset based on the number of inlier matches of multiple Homography-RANSAC.

\section{Experiments}
\label{sec:exp}
We implement feature detection, description and matching (Sections~\ref{sec:denseextract} to~\ref{sec:ransac}) in MATLAB with third-party libraries (MatConvNet~\cite{vedaldi15matconvnet} and Yael library~\cite{Douze:2014:YL:2647868.2654892}).
Dense CNN features are extracted using the VGG-16 network~\cite{simonyan2014very}. Using conv4 and conv3 max-pooling layers, feature matches are computed by the coarse-to-fine matching followed by multiple Homography-RANSAC that finds at most five homographies supported by an inlier threshold of 10 pixels.
The best $N$ pairs based on multiple Homography-RANSAC of every image are imported to COLMAP~\cite{schonberger_structure--motion_2016} with the fixed intrinsic parameter option for scene with many repetitive structures.
Otherwise, we use all the image pairs.

In our preliminary experiments, we tested other layers having the same spatial resolution, \eg, using conv4\_3 and conv3\_3 layers in the coarse-to-fine matching but we observed no improvement in 3D reconstruction.
As a max-pooling layer has a half depth dimension in comparison with the other layers at the same spatial resolution, we chose the max-pooling layer as the dense features for efficiency. 

In the following, we evaluate the reconstruction performance on Aachen Day-Night~\cite{sattler2017benchmarking} and Strecha~\cite{strecha_benchmarking_2008} dataset. 
We compare our SfM using dense CNN features with keypoint relocalization to the baseline COLMAP with DoG+RootSIFT features~\cite{schonberger_structure--motion_2016}.
In addition, we also compare our SfM to SfM using dense CNN without keypoint relocalization~\cite{sattler2017benchmarking}.
All experiments are tested on a computer equipped with a 3.20GHz Intel Core i7-6900K CPU with 16 threads and a 12GB GeForce GTX 1080Ti.

\begin{figure*}[tb]
\centering
\subfloat[]{
   \includegraphics[width=0.95\linewidth]{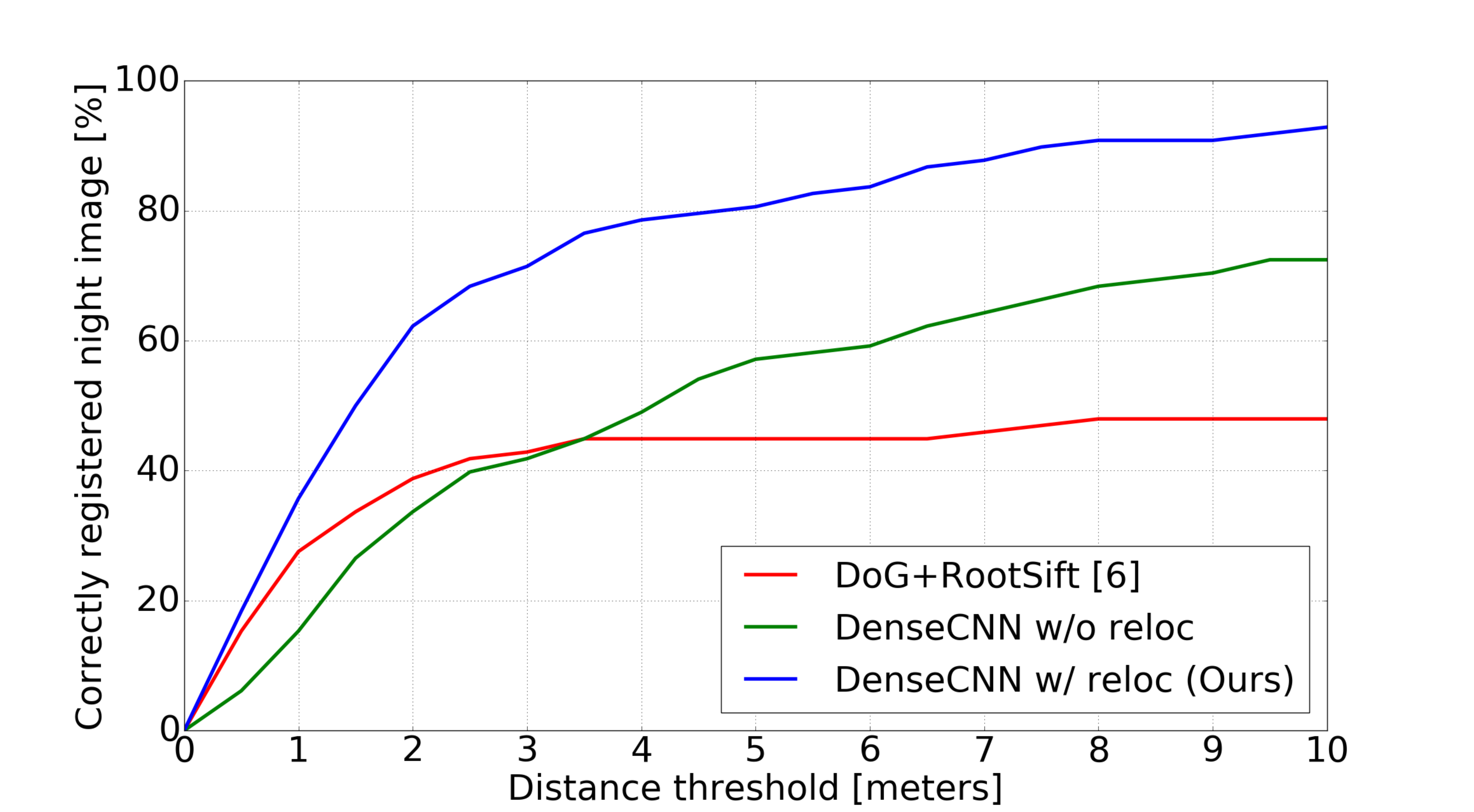}
 }
\subfloat[]{
   \includegraphics[width=0.95\linewidth]{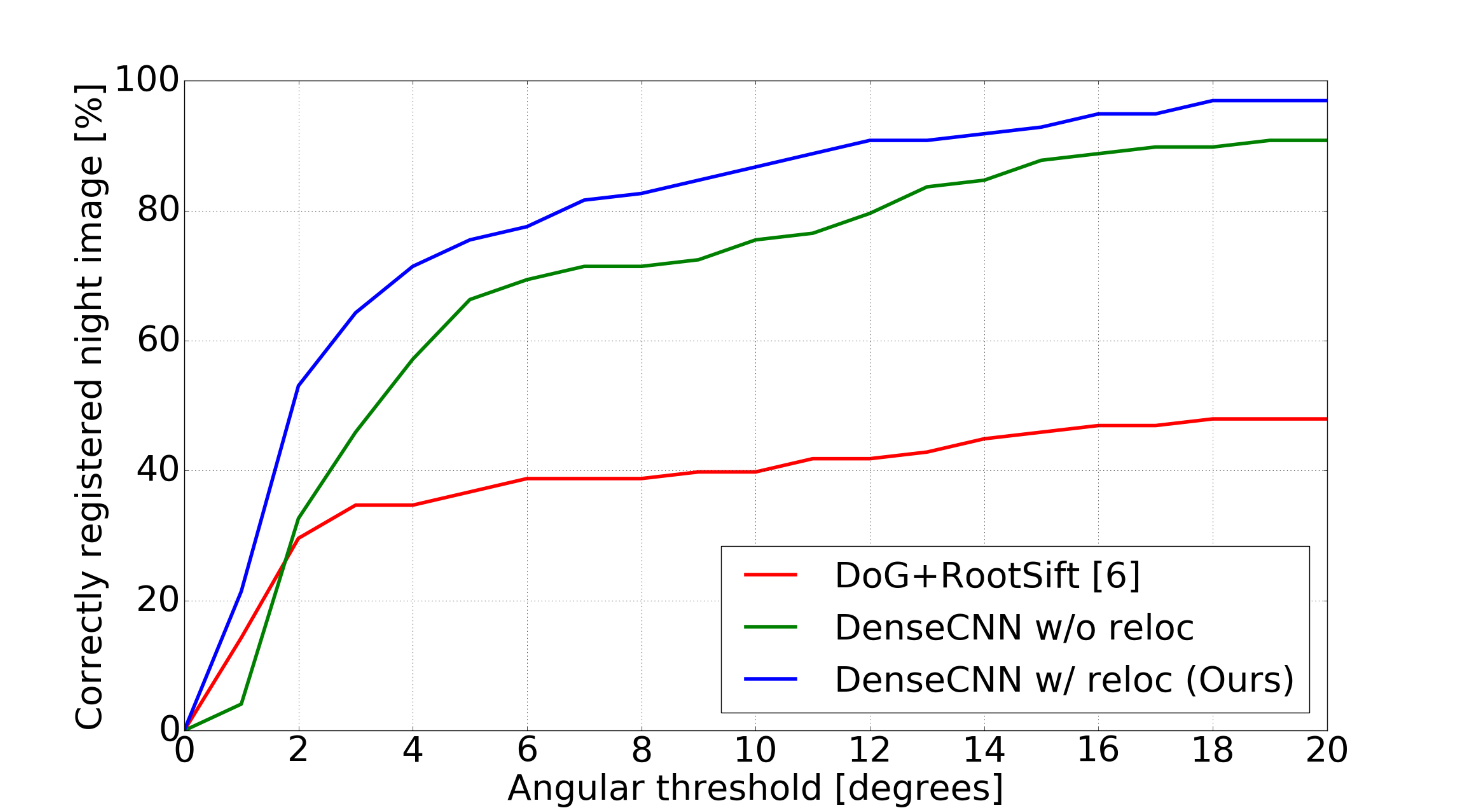}
 }
\caption{{\bf Quantitative evaluation on the Aachen Day-Night dataset}. The poses of night images reconstructed by the baseline DoG+RootSIFT~\cite{schonberger_structure--motion_2016} (red), the DenseCNN without keypoint relocalization (green), and the proposed DenseCNN (blue) are evaluated using the reference poses. The graphs show the percentages of correctly reconstructed camera poses of night images (y-axis) at positional (a) and angular (b) error threshold (x-axis). 
}
\label{fig:aachen-pos}
\end{figure*}

\subsection{Results on Aachen Day-Night dataset}
The Aachen Day-Night dataset~\cite{sattler2017benchmarking} is aimed for evaluating SfM and visual localization under large illumination changes such as day and night. It includes 98 subsets of images. Each subset consists of 20 day-time images and one night-time image, their reference camera poses, and 3D points~\footnote{Although the poses are carefully obtained with manual verification, the poses are called as ``reference poses'' but not ground truth.}.

For each subset, we run SfM and evaluate the estimated camera pose of the night image as follows. First, 
the reconstructed SfM model is registered to the reference camera poses by adopting a similarity transform obtained from the camera positions of day-time images.
We then evaluate the estimated camera pose of the night image by measuring positional (L2 distance) and angular ($\text{acos}(\frac{\text{trace}(\boldsymbol{R}_{ref}\boldsymbol{R}_{night}^T)-1}{2})$) error.

Table~\ref{tbl:aachen} shows the number of reconstructed cameras. The proposed SfM with keypoint relocalization (conv1\_2) can reconstruct 96 night images that are twice as many as that of the baseline method using COLMAP with DoG+RootSIFT~\cite{schonberger_structure--motion_2016}. This result validates the benefit of densely detected features that can provide correspondences across large illumination changes as they have smaller loss in keypoint detection repeatablity than a standard DoG. On the other hand, both methods with sparse and dense features work well for reconstructing day images. The difference between with and without keypoint localization can be seen more clearly in the next evaluation. 
\begin{table}[tb]
\centering
\caption{Number of cameras reconstructed on the Aachen dataset}
\label{tbl:aachen}
\begin{tabular}{c|cccc}
& DoG+     & DenseCNN  & DenseCNN  \\ 
& RootSIFT~\cite{schonberger_structure--motion_2016} & w/o reloc & w/ reloc (Ours) \\ \hline
Night & 48 & 95 & \textbf{96} \\ 
Day   & 1910 & 1924 & \textbf{1944} \\
\end{tabular}
\end{table}

\begin{figure*}[tb]
\centering
\subfloat[]{
   \includegraphics[width=0.9\linewidth]{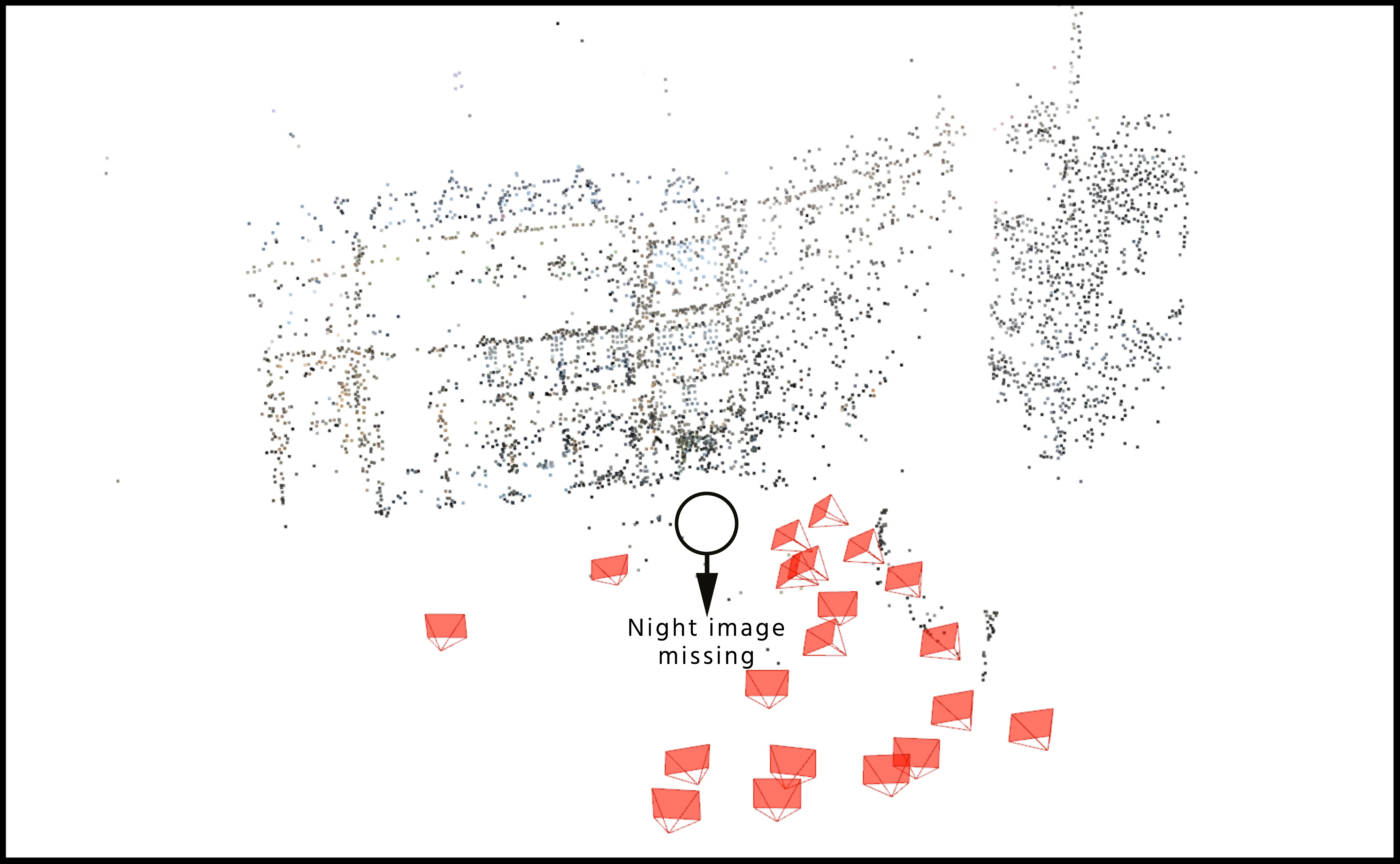}   
 }
\subfloat[]{
   \includegraphics[width=0.9\linewidth]{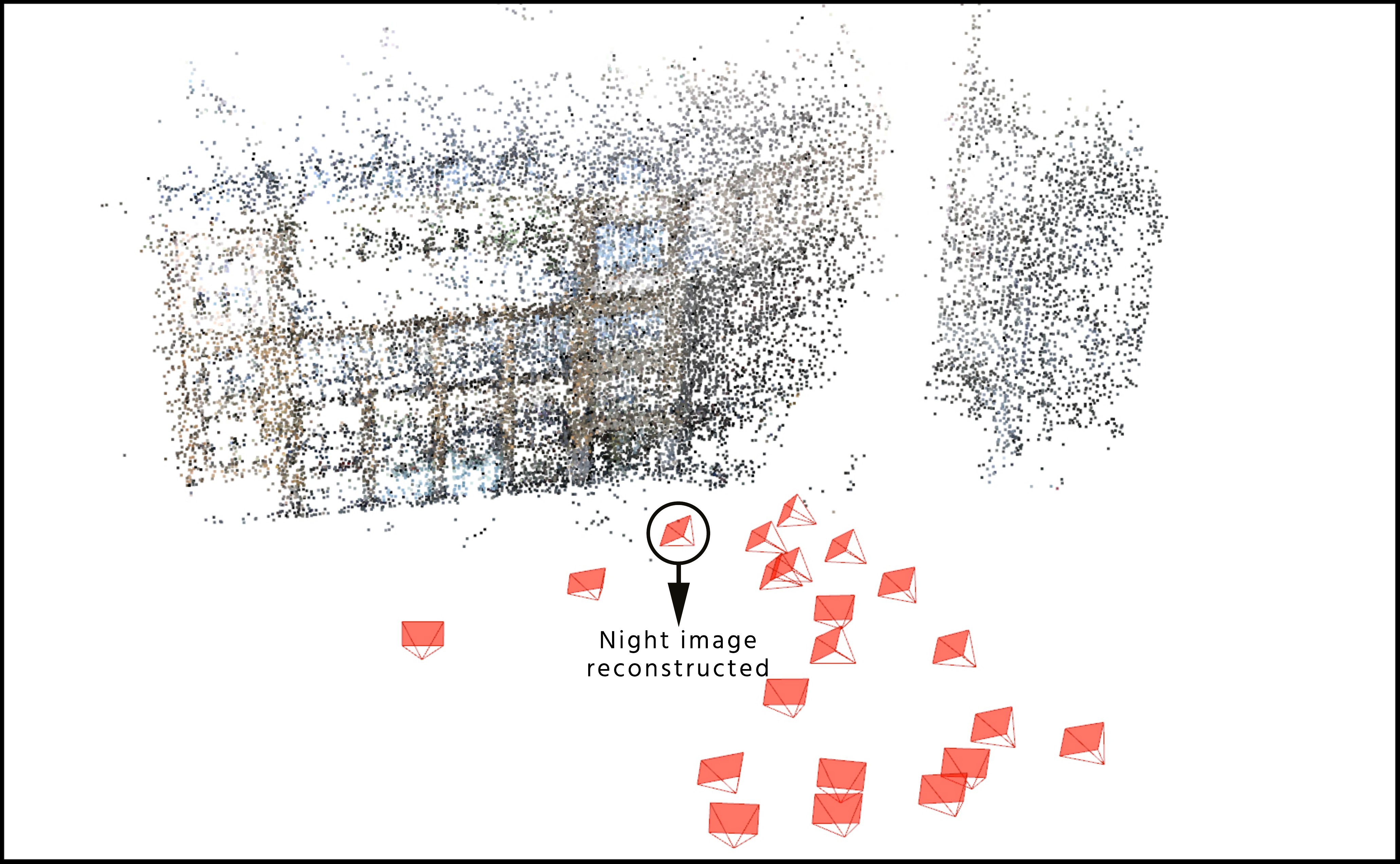}
 }
\caption{{\bf Example of 3D reconstruction in the Aachen dataset}. These figures show qualitative examples of SfM using DoG+RootSIFT~\cite{schonberger_structure--motion_2016} (a) and our dense CNN with keypoint relocalization (b). Our method can reconstruct all the 21 images in the subset whereas the baseline DoG+RootSIFT~\cite{schonberger_structure--motion_2016} fails to reconstruct it. As a nature of dense feature matching, our method reconstructs 42,402 3D points which are $8.2$ times more than the baseline method. 
}
\label{fig:proposed-aachen}
\end{figure*}

Figure~\ref{fig:aachen-pos} shows the percentages of night images reconstructed (y-axis) within certain positional and angular error threshold (x-axis). Similarly, Table~\ref{tbl:aachen-pos} shows the reconstruction percentages of night images for varying distance error thresholds with a fixed angular error threshold at ten degrees. 
As can be seen from both evaluations, the proposed SfM using dense CNN features with keypoint relocalization outperforms the baseline DoG+RootSIFT~\cite{schonberger_structure--motion_2016} by a large margin. 
The improvement by the proposed keypoint relocalization is significant when the evaluation accounts for pose accuracy.
Notice that the SfM using dense CNN without keypoint relocalization~\cite{sattler2017benchmarking} performs worse than the baseline DoG+RootSIFT~\cite{schonberger_structure--motion_2016} at small thresholds, \eg, below $3.5$ meters position and $2^o$ angular error. This indicates that the proposed keypoint relocalization gives features at more stable and accurate positions and provides better inlier matches for COLMAP reconstruction which results 3D reconstruction in higher quality.

Figure~\ref{fig:proposed-aachen} illustrates the qualitative comparison result between our method and the baseline DoG+Root\-SIFT~\cite{schonberger_structure--motion_2016}.

\begin{table}[tb]
\centering
\caption{Evaluation of reconstructed camera poses (both position and orientation). The numbers show the percentage of the reconstructed night images within given positional error thresholds and an angular error fixed at $10^o$.}
\label{tbl:aachen-pos}
\begin{tabular}{r|ccc}
& DoG+     & DenseCNN  & DenseCNN  \\ 
& RootSIFT~\cite{schonberger_structure--motion_2016} & w/o reloc & w/ reloc (Ours) \\ \hline
0.5m & 15.31 & 5.10 & \textbf{18.37} \\
1.0m  & 25.61 & 14.29  & \textbf{33.67}   \\
5.0m & 36.73 & 45.92  & \textbf{69.39}   \\\
10.0m & 35.71 & 61.22  & \textbf{81.63}   \\ 
20.0m & 39.80 & 69.39 & \textbf{82.65} \\ \hline
\end{tabular}
\end{table}

\subsection{Results on Strecha dataset}
We additionally evaluate our SfM using dense CNN with the proposed keypoint relocalization on all six subsets of Strecha dataset~\cite{strecha_benchmarking_2008} which is a standard benchmark dataset for SfM and MVS.
Position and angular error between the reconstructed cameras and the ground truth poses are evaluated.
In our SfM, we take only feature matches from the best $N=5$ image pairs for each image to suppress artifacts from confusing image pairs. 

The mean average position and angular errors resulted by our SfM are $0.59$ meters and $2.27$ degrees. Although these errors are worse than those of the state-of-the-art COLMAP with DoG+RootSIFT~\cite{schonberger_structure--motion_2016} which are $0.17$ meters and $0.90$ degrees, the quantitative evaluation on the Strecha dataset demonstrated that our SfM does not overfit to specific challenging tasks but works reasonably well for standard (easy) situations. 

\section{Conclusion}
We presented a new SfM using dense features extracted from CNN with the proposed keypoint relocalization to improve the accuracy of feature positions sampled on a regular grid. The advantage of our SfM has demonstrated on the Aachen Day-Night dataset that includes images with large illumination changes. The result on the Strecha dataset also showed that our SfM works for standard datasets and does not overfit to a particular task although it is less accurate than the state-of-the-art SfM with local features. We wish the proposed SfM becomes a milestone in the 3D reconstruction, in particularly, challenging situations. 



\bibliographystyle{bmc-mathphys} 

  \newcommand{\BMCxmlcomment}[1]{}
  
  \BMCxmlcomment{
  
  <refgrp>
  
  <bibl id="B1">
    <title><p>Pix4D - Professional drone mapping and photogrammetry
    software</p></title>
    <url>https://pix4d.com/</url>
  </bibl>
  
  <bibl id="B2">
    <title><p>Agisoft Photoscan</p></title>
    <url>http://www.agisoft.com/</url>
  </bibl>
  
  <bibl id="B3">
    <title><p>Discover Photogrammetry Software - Photomodeler</p></title>
    <url>http://www.photomodeler.com/index.html</url>
  </bibl>
  
  <bibl id="B4">
    <title><p>{MVE}-{A} {Multi}-{View} {Reconstruction}
    {Environment}.</p></title>
    <aug>
      <au><snm>Fuhrmann</snm><fnm>S</fnm></au>
      <au><snm>Langguth</snm><fnm>F</fnm></au>
      <au><snm>Goesele</snm><fnm>M</fnm></au>
    </aug>
    <source>{GCH}</source>
    <pubdate>2014</pubdate>
    <fpage>11</fpage>
    <lpage>-18</lpage>
  </bibl>
  
  <bibl id="B5">
    <title><p>Theia: {A} fast and scalable structure-from-motion
    library</p></title>
    <aug>
      <au><snm>Sweeney</snm><fnm>C</fnm></au>
      <au><snm>Hollerer</snm><fnm>T</fnm></au>
      <au><snm>Turk</snm><fnm>M</fnm></au>
    </aug>
    <source>Proc. ACMM</source>
    <pubdate>2015</pubdate>
    <fpage>693</fpage>
    <lpage>-696</lpage>
  </bibl>
  
  <bibl id="B6">
    <title><p>Structure-from-motion revisited</p></title>
    <aug>
      <au><snm>Schonberger</snm><fnm>JL</fnm></au>
      <au><snm>Frahm</snm><fnm>JM</fnm></au>
    </aug>
    <source>Proc. CVPR</source>
    <pubdate>2016</pubdate>
    <fpage>4104</fpage>
    <lpage>-4113</lpage>
  </bibl>
  
  <bibl id="B7">
    <title><p>Pixelwise view selection for unstructured multi-view
    stereo</p></title>
    <aug>
      <au><snm>Schönberger</snm><fnm>JL</fnm></au>
      <au><snm>Zheng</snm><fnm>E</fnm></au>
      <au><snm>Frahm</snm><fnm>JM</fnm></au>
      <au><snm>Pollefeys</snm><fnm>M</fnm></au>
    </aug>
    <source>Proc. ECCV</source>
    <pubdate>2016</pubdate>
    <fpage>501</fpage>
    <lpage>-518</lpage>
  </bibl>
  
  <bibl id="B8">
    <title><p>Robust global translations with 1dsfm</p></title>
    <aug>
      <au><snm>Wilson</snm><fnm>K</fnm></au>
      <au><snm>Snavely</snm><fnm>N</fnm></au>
    </aug>
    <source>Proc. ECCV</source>
    <pubdate>2014</pubdate>
    <fpage>61</fpage>
    <lpage>-75</lpage>
  </bibl>
  
  <bibl id="B9">
    <title><p>{OpenMVG}: {Open} multiple view geometry</p></title>
    <aug>
      <au><snm>Moulon</snm><fnm>P</fnm></au>
      <au><snm>Monasse</snm><fnm>P</fnm></au>
      <au><snm>Perrot</snm><fnm>R</fnm></au>
      <au><snm>Marlet</snm><fnm>R</fnm></au>
    </aug>
    <source>International {Workshop} on {Reproducible} {Research} in {Pattern}
    {Recognition}</source>
    <pubdate>2016</pubdate>
    <fpage>60</fpage>
    <lpage>-74</lpage>
  </bibl>
  
  <bibl id="B10">
    <title><p>HSfM: Hybrid Structure-from-Motion</p></title>
    <aug>
      <au><snm>Cui</snm><fnm>H</fnm></au>
      <au><snm>Gao</snm><fnm>X</fnm></au>
      <au><snm>Shen</snm><fnm>S</fnm></au>
      <au><snm>Hu</snm><fnm>Z</fnm></au>
    </aug>
  </bibl>
  
  <bibl id="B11">
    <title><p>Distinctive image features from scale-invariant
    keypoints</p></title>
    <aug>
      <au><snm>Lowe</snm><fnm>DG</fnm></au>
    </aug>
    <source>IJCV</source>
    <pubdate>2004</pubdate>
    <volume>60</volume>
    <issue>2</issue>
    <fpage>91</fpage>
    <lpage>-110</lpage>
  </bibl>
  
  <bibl id="B12">
    <title><p>Scale &amp affine invariant interest point detectors</p></title>
    <aug>
      <au><snm>Mikolajczyk</snm><fnm>K</fnm></au>
      <au><snm>Schmid</snm><fnm>C</fnm></au>
    </aug>
    <source>IJCV</source>
    <publisher>Springer</publisher>
    <pubdate>2004</pubdate>
    <volume>60</volume>
    <issue>1</issue>
    <fpage>63</fpage>
    <lpage>-86</lpage>
  </bibl>
  
  <bibl id="B13">
    <title><p>An affine invariant salient region detector</p></title>
    <aug>
      <au><snm>Kadir</snm><fnm>T</fnm></au>
      <au><snm>Zisserman</snm><fnm>A</fnm></au>
      <au><snm>Brady</snm><fnm>M</fnm></au>
    </aug>
    <source>Proc. ECCV</source>
    <pubdate>2004</pubdate>
    <fpage>228</fpage>
    <lpage>-241</lpage>
  </bibl>
  
  <bibl id="B14">
    <title><p>Matching widely separated views based on affine invariant
    regions</p></title>
    <aug>
      <au><snm>Tuytelaars</snm><fnm>T</fnm></au>
      <au><snm>Van Gool</snm><fnm>L</fnm></au>
    </aug>
    <source>IJCV</source>
    <publisher>Springer</publisher>
    <pubdate>2004</pubdate>
    <volume>59</volume>
    <issue>1</issue>
    <fpage>61</fpage>
    <lpage>-85</lpage>
  </bibl>
  
  <bibl id="B15">
    <title><p>Three things everyone should know to improve object
    retrieval</p></title>
    <aug>
      <au><snm>Arandjelovi{\'c}</snm><fnm>R</fnm></au>
      <au><snm>Zisserman</snm><fnm>A</fnm></au>
    </aug>
    <source>Proc. CVPR</source>
    <pubdate>2012</pubdate>
    <fpage>2911</fpage>
    <lpage>-2918</lpage>
  </bibl>
  
  <bibl id="B16">
    <title><p>Domain-size pooling in local descriptors: DSP-SIFT</p></title>
    <aug>
      <au><snm>Dong</snm><fnm>J</fnm></au>
      <au><snm>Soatto</snm><fnm>S</fnm></au>
    </aug>
    <source>Proc. CVPR</source>
    <pubdate>2015</pubdate>
    <fpage>5097</fpage>
    <lpage>-5106</lpage>
  </bibl>
  
  <bibl id="B17">
    <title><p>Lift: {Learned} invariant feature transform</p></title>
    <aug>
      <au><snm>Yi</snm><fnm>KM</fnm></au>
      <au><snm>Trulls</snm><fnm>E</fnm></au>
      <au><snm>Lepetit</snm><fnm>V</fnm></au>
      <au><snm>Fua</snm><fnm>P</fnm></au>
    </aug>
    <source>Proc. ECCV</source>
    <pubdate>2016</pubdate>
    <fpage>467</fpage>
    <lpage>-483</lpage>
  </bibl>
  
  <bibl id="B18">
    <title><p>Modeling the world from internet photo collections</p></title>
    <aug>
      <au><snm>Snavely</snm><fnm>N</fnm></au>
      <au><snm>Seitz</snm><fnm>SM</fnm></au>
      <au><snm>Szeliski</snm><fnm>R</fnm></au>
    </aug>
    <source>IJCV</source>
    <pubdate>2008</pubdate>
    <volume>80</volume>
    <issue>2</issue>
    <fpage>189</fpage>
    <lpage>-210</lpage>
  </bibl>
  
  <bibl id="B19">
    <title><p>Reconstructing rome</p></title>
    <aug>
      <au><snm>Agarwal</snm><fnm>S</fnm></au>
      <au><snm>Furukawa</snm><fnm>Y</fnm></au>
      <au><snm>Snavely</snm><fnm>N</fnm></au>
      <au><snm>Curless</snm><fnm>B</fnm></au>
      <au><snm>Seitz</snm><fnm>SM</fnm></au>
      <au><snm>Szeliski</snm><fnm>R</fnm></au>
    </aug>
    <source>Computer</source>
    <pubdate>2010</pubdate>
    <volume>43</volume>
    <issue>6</issue>
    <fpage>40</fpage>
    <lpage>-47</lpage>
  </bibl>
  
  <bibl id="B20">
    <title><p>Building rome in a day</p></title>
    <aug>
      <au><snm>Agarwal</snm><fnm>S</fnm></au>
      <au><snm>Furukawa</snm><fnm>Y</fnm></au>
      <au><snm>Snavely</snm><fnm>N</fnm></au>
      <au><snm>Simon</snm><fnm>I</fnm></au>
      <au><snm>Curless</snm><fnm>B</fnm></au>
      <au><snm>Seitz</snm><fnm>SM</fnm></au>
      <au><snm>Szeliski</snm><fnm>R</fnm></au>
    </aug>
    <source>Communications of the ACM</source>
    <pubdate>2011</pubdate>
    <volume>54</volume>
    <issue>10</issue>
    <fpage>105</fpage>
    <lpage>-112</lpage>
  </bibl>
  
  <bibl id="B21">
    <title><p>ASIFT: A new framework for fully affine invariant image
    comparison</p></title>
    <aug>
      <au><snm>Morel</snm><fnm>JM</fnm></au>
      <au><snm>Yu</snm><fnm>G</fnm></au>
    </aug>
    <source>SIAM journal on imaging sciences</source>
    <publisher>SIAM</publisher>
    <pubdate>2009</pubdate>
    <volume>2</volume>
    <issue>2</issue>
    <fpage>438</fpage>
    <lpage>-469</lpage>
  </bibl>
  
  <bibl id="B22">
    <title><p>PCA-SIFT: A more distinctive representation for local image
    descriptors</p></title>
    <aug>
      <au><snm>Ke</snm><fnm>Y</fnm></au>
      <au><snm>Sukthankar</snm><fnm>R</fnm></au>
    </aug>
    <source>Proc. CVPR</source>
    <pubdate>2004</pubdate>
    <volume>2</volume>
    <fpage>II</fpage>
    <lpage>-II</lpage>
  </bibl>
  
  <bibl id="B23">
    <title><p>CSIFT: A SIFT descriptor with color invariant
    characteristics</p></title>
    <aug>
      <au><snm>Abdel Hakim</snm><fnm>AE</fnm></au>
      <au><snm>Farag</snm><fnm>AA</fnm></au>
    </aug>
    <source>Proc. CVPR</source>
    <pubdate>2006</pubdate>
    <volume>2</volume>
    <fpage>1978</fpage>
    <lpage>-1983</lpage>
  </bibl>
  
  <bibl id="B24">
    <title><p>Surf: Speeded up robust features</p></title>
    <aug>
      <au><snm>Bay</snm><fnm>H</fnm></au>
      <au><snm>Tuytelaars</snm><fnm>T</fnm></au>
      <au><snm>Van Gool</snm><fnm>L</fnm></au>
    </aug>
    <source>Proc. ECCV</source>
    <pubdate>2006</pubdate>
    <fpage>404</fpage>
    <lpage>-417</lpage>
  </bibl>
  
  <bibl id="B25">
    <title><p>Visual categorization with bags of keypoints</p></title>
    <aug>
      <au><snm>Csurka</snm><fnm>G</fnm></au>
      <au><snm>Dance</snm><fnm>C</fnm></au>
      <au><snm>Fan</snm><fnm>L</fnm></au>
      <au><snm>Willamowski</snm><fnm>J</fnm></au>
      <au><snm>Bray</snm><fnm>C</fnm></au>
    </aug>
    <source>Proc. ECCV</source>
    <pubdate>2004</pubdate>
    <volume>1</volume>
    <issue>1-22</issue>
    <fpage>1</fpage>
    <lpage>-2</lpage>
  </bibl>
  
  <bibl id="B26">
    <title><p>Beyond bags of features: Spatial pyramid matching for recognizing
    natural scene categories</p></title>
    <aug>
      <au><snm>Lazebnik</snm><fnm>S</fnm></au>
      <au><snm>Schmid</snm><fnm>C</fnm></au>
      <au><snm>Ponce</snm><fnm>J</fnm></au>
    </aug>
    <source>Proc. CVPR</source>
    <pubdate>2006</pubdate>
    <volume>2</volume>
    <fpage>2169</fpage>
    <lpage>-2178</lpage>
  </bibl>
  
  <bibl id="B27">
    <title><p>Simultaneous image classification and annotation</p></title>
    <aug>
      <au><snm>Chong</snm><fnm>W</fnm></au>
      <au><snm>Blei</snm><fnm>D</fnm></au>
      <au><snm>Li</snm><fnm>FF</fnm></au>
    </aug>
    <source>Computer Vision and Pattern Recognition, 2009. CVPR 2009. IEEE
    Conference on</source>
    <pubdate>2009</pubdate>
    <fpage>1903</fpage>
    <lpage>-1910</lpage>
  </bibl>
  
  <bibl id="B28">
    <title><p>Gradient response maps for real-time detection of textureless
    objects</p></title>
    <aug>
      <au><snm>Hinterstoisser</snm><fnm>S</fnm></au>
      <au><snm>Cagniart</snm><fnm>C</fnm></au>
      <au><snm>Ilic</snm><fnm>S</fnm></au>
      <au><snm>Sturm</snm><fnm>P</fnm></au>
      <au><snm>Navab</snm><fnm>N</fnm></au>
      <au><snm>Fua</snm><fnm>P</fnm></au>
      <au><snm>Lepetit</snm><fnm>V</fnm></au>
    </aug>
    <source>IEEE PAMI</source>
    <publisher>IEEE</publisher>
    <pubdate>2012</pubdate>
    <volume>34</volume>
    <issue>5</issue>
    <fpage>876</fpage>
    <lpage>-888</lpage>
  </bibl>
  
  <bibl id="B29">
    <title><p>Visual place recognition with repetitive structures</p></title>
    <aug>
      <au><snm>Torii</snm><fnm>A</fnm></au>
      <au><snm>Sivic</snm><fnm>J</fnm></au>
      <au><snm>Pajdla</snm><fnm>T</fnm></au>
      <au><snm>Okutomi</snm><fnm>M</fnm></au>
    </aug>
    <source>Proc. CVPR</source>
    <pubdate>2013</pubdate>
    <fpage>883</fpage>
    <lpage>-890</lpage>
  </bibl>
  
  <bibl id="B30">
    <title><p>Mods: Fast and robust method for two-view matching</p></title>
    <aug>
      <au><snm>Mishkin</snm><fnm>D</fnm></au>
      <au><snm>Matas</snm><fnm>J</fnm></au>
      <au><snm>Perdoch</snm><fnm>M</fnm></au>
    </aug>
    <source>CVIU</source>
    <publisher>Elsevier</publisher>
    <pubdate>2015</pubdate>
    <volume>141</volume>
    <fpage>81</fpage>
    <lpage>-93</lpage>
  </bibl>
  
  <bibl id="B31">
    <title><p>Robust feature matching by learning descriptor covariance with
    viewpoint synthesis</p></title>
    <aug>
      <au><snm>Taira</snm><fnm>H</fnm></au>
      <au><snm>Torii</snm><fnm>A</fnm></au>
      <au><snm>Okutomi</snm><fnm>M</fnm></au>
    </aug>
    <source>Proc. ICPR</source>
    <pubdate>2016</pubdate>
    <fpage>1953</fpage>
    <lpage>-1958</lpage>
  </bibl>
  
  <bibl id="B32">
    <title><p>24/7 place recognition by view synthesis</p></title>
    <aug>
      <au><snm>Torii</snm><fnm>A</fnm></au>
      <au><snm>Arandjelović</snm><fnm>R</fnm></au>
      <au><snm>Sivic</snm><fnm>J</fnm></au>
      <au><snm>Okutomi</snm><fnm>M</fnm></au>
      <au><snm>Pajdla</snm><fnm>T</fnm></au>
    </aug>
    <source>Proc. CVPR</source>
    <pubdate>2015</pubdate>
    <fpage>1808</fpage>
    <lpage>-1817</lpage>
  </bibl>
  
  <bibl id="B33">
    <title><p>From dusk till dawn: {Modeling} in the dark</p></title>
    <aug>
      <au><snm>Radenovic</snm><fnm>F</fnm></au>
      <au><snm>Schonberger</snm><fnm>JL</fnm></au>
      <au><snm>Ji</snm><fnm>D</fnm></au>
      <au><snm>Frahm</snm><fnm>JM</fnm></au>
      <au><snm>Chum</snm><fnm>O</fnm></au>
      <au><snm>Matas</snm><fnm>J</fnm></au>
    </aug>
    <source>Proc. CVPR</source>
    <pubdate>2016</pubdate>
    <fpage>5488</fpage>
    <lpage>-5496</lpage>
  </bibl>
  
  <bibl id="B34">
    <title><p>Image classification using random forests and ferns</p></title>
    <aug>
      <au><snm>Bosch</snm><fnm>A</fnm></au>
      <au><snm>Zisserman</snm><fnm>A</fnm></au>
      <au><snm>Munoz</snm><fnm>X</fnm></au>
    </aug>
    <source>Proc. ICCV</source>
    <pubdate>2007</pubdate>
    <fpage>1</fpage>
    <lpage>-8</lpage>
  </bibl>
  
  <bibl id="B35">
    <title><p>Sift flow: Dense correspondence across scenes and its
    applications</p></title>
    <aug>
      <au><snm>Liu</snm><fnm>C</fnm></au>
      <au><snm>Yuen</snm><fnm>J</fnm></au>
      <au><snm>Torralba</snm><fnm>A</fnm></au>
    </aug>
    <source>Dense Image Correspondences for Computer Vision</source>
    <pubdate>2016</pubdate>
    <fpage>15</fpage>
    <lpage>-49</lpage>
  </bibl>
  
  <bibl id="B36">
    <title><p>Oriented pooling for dense and non-dense rotation-invariant
    features</p></title>
    <aug>
      <au><snm>Zhao</snm><fnm>WL</fnm></au>
      <au><snm>J{\'e}gou</snm><fnm>H</fnm></au>
      <au><snm>Gravier</snm><fnm>G</fnm></au>
    </aug>
    <source>Proc. BMVC.</source>
    <pubdate>2013</pubdate>
  </bibl>
  
  <bibl id="B37">
    <title><p>Benchmarking 6DOF Outdoor Visual Localization in Changing
    Conditions</p></title>
    <aug>
      <au><snm>Sattler</snm><fnm>T</fnm></au>
      <au><snm>Maddern</snm><fnm>W</fnm></au>
      <au><snm>Toft</snm><fnm>C</fnm></au>
      <au><snm>Torii</snm><fnm>A</fnm></au>
      <au><snm>Hammarstrand</snm><fnm>L</fnm></au>
      <au><snm>Stenborg</snm><fnm>E</fnm></au>
      <au><snm>Safari</snm><fnm>D</fnm></au>
      <au><snm>Sivic</snm><fnm>J</fnm></au>
      <au><snm>Pajdla</snm><fnm>T</fnm></au>
      <au><snm>Pollefeys</snm><fnm>M</fnm></au>
      <au><snm>Kahl</snm><fnm>F</fnm></au>
      <au><snm>Okutomi</snm><fnm>M</fnm></au>
    </aug>
    <source>arXiv preprint arXiv:1707.09092</source>
    <pubdate>2017</pubdate>
  </bibl>
  
  <bibl id="B38">
    <title><p>Very deep convolutional networks for large-scale image
    recognition</p></title>
    <aug>
      <au><snm>Simonyan</snm><fnm>K</fnm></au>
      <au><snm>Zisserman</snm><fnm>A</fnm></au>
    </aug>
    <source>arXiv preprint arXiv:1409.1556</source>
    <pubdate>2014</pubdate>
  </bibl>
  
  <bibl id="B39">
    <title><p>On benchmarking camera calibration and multi-view stereo for high
    resolution imagery</p></title>
    <aug>
      <au><snm>Strecha</snm><fnm>C</fnm></au>
      <au><snm>Von Hansen</snm><fnm>W</fnm></au>
      <au><snm>Van Gool</snm><fnm>L</fnm></au>
      <au><snm>Fua</snm><fnm>P</fnm></au>
      <au><snm>Thoennessen</snm><fnm>U</fnm></au>
    </aug>
    <source>Proc. CVPR</source>
    <pubdate>2008</pubdate>
    <fpage>1</fpage>
    <lpage>-8</lpage>
  </bibl>
  
  <bibl id="B40">
    <title><p>Towards linear-time incremental structure from motion</p></title>
    <aug>
      <au><snm>Wu</snm><fnm>C</fnm></au>
    </aug>
    <source>Proc. 3DV</source>
    <pubdate>2013</pubdate>
    <fpage>127</fpage>
    <lpage>-134</lpage>
  </bibl>
  
  <bibl id="B41">
    <title><p>Global structure-from-motion by similarity averaging</p></title>
    <aug>
      <au><snm>Cui</snm><fnm>Z</fnm></au>
      <au><snm>Tan</snm><fnm>P</fnm></au>
    </aug>
    <source>Proceedings of the IEEE International Conference on Computer
    Vision</source>
    <pubdate>2015</pubdate>
    <fpage>864</fpage>
    <lpage>-872</lpage>
  </bibl>
  
  <bibl id="B42">
    <title><p>Practical Projective Structure from Motion (P2SfM)</p></title>
    <aug>
      <au><snm>Magerand</snm><fnm>L</fnm></au>
      <au><snm>Del Bue</snm><fnm>A</fnm></au>
    </aug>
    <source>Proc. CVPR</source>
    <pubdate>2017</pubdate>
    <fpage>39</fpage>
    <lpage>-47</lpage>
  </bibl>
  
  <bibl id="B43">
    <title><p>LSD-SLAM: Large-scale direct monocular SLAM</p></title>
    <aug>
      <au><snm>Engel</snm><fnm>J</fnm></au>
      <au><snm>Sch{\"o}ps</snm><fnm>T</fnm></au>
      <au><snm>Cremers</snm><fnm>D</fnm></au>
    </aug>
    <source>Proc. ECCV</source>
    <pubdate>2014</pubdate>
    <fpage>834</fpage>
    <lpage>-849</lpage>
  </bibl>
  
  <bibl id="B44">
    <title><p>DTAM: Dense tracking and mapping in real-time</p></title>
    <aug>
      <au><snm>Newcombe</snm><fnm>RA</fnm></au>
      <au><snm>Lovegrove</snm><fnm>SJ</fnm></au>
      <au><snm>Davison</snm><fnm>AJ</fnm></au>
    </aug>
    <source>Proc. ICCV</source>
    <pubdate>2011</pubdate>
    <fpage>2320</fpage>
    <lpage>-2327</lpage>
  </bibl>
  
  <bibl id="B45">
    <title><p>Multi-view stereo: A tutorial</p></title>
    <aug>
      <au><snm>Furukawa</snm><fnm>Y</fnm></au>
      <au><snm>Hern{\'a}ndez</snm><fnm>C</fnm></au>
      <au><cnm>others</cnm></au>
    </aug>
    <source>Foundations and Trends{\textregistered} in Computer Graphics and
    Vision</source>
    <publisher>Now Publishers, Inc.</publisher>
    <pubdate>2015</pubdate>
    <volume>9</volume>
    <issue>1-2</issue>
    <fpage>1</fpage>
    <lpage>-148</lpage>
  </bibl>
  
  <bibl id="B46">
    <title><p>Daisy: {An} efficient dense descriptor applied to wide-baseline
    stereo</p></title>
    <aug>
      <au><snm>Tola</snm><fnm>E</fnm></au>
      <au><snm>Lepetit</snm><fnm>V</fnm></au>
      <au><snm>Fua</snm><fnm>P</fnm></au>
    </aug>
    <source>IEEE PAMI</source>
    <pubdate>2010</pubdate>
    <volume>32</volume>
    <issue>5</issue>
    <fpage>815</fpage>
    <lpage>-830</lpage>
  </bibl>
  
  <bibl id="B47">
    <title><p>Dense interest points</p></title>
    <aug>
      <au><snm>Tuytelaars</snm><fnm>T</fnm></au>
    </aug>
    <source>Proc. CVPR</source>
    <pubdate>2010</pubdate>
    <fpage>2281</fpage>
    <lpage>-2288</lpage>
  </bibl>
  
  <bibl id="B48">
    <title><p>Descriptor matching with convolutional neural networks: a
    comparison to sift</p></title>
    <aug>
      <au><snm>Fischer</snm><fnm>P</fnm></au>
      <au><snm>Dosovitskiy</snm><fnm>A</fnm></au>
      <au><snm>Brox</snm><fnm>T</fnm></au>
    </aug>
    <source>arXiv preprint arXiv:1405.5769</source>
    <pubdate>2014</pubdate>
  </bibl>
  
  <bibl id="B49">
    <title><p>Comparative evaluation of hand-crafted and learned local
    features</p></title>
    <aug>
      <au><snm>Schonberger</snm><fnm>JL</fnm></au>
      <au><snm>Hardmeier</snm><fnm>H</fnm></au>
      <au><snm>Sattler</snm><fnm>T</fnm></au>
      <au><snm>Pollefeys</snm><fnm>M</fnm></au>
    </aug>
    <source>Proc. CVPR</source>
    <pubdate>2017</pubdate>
    <fpage>6959</fpage>
    <lpage>-6968</lpage>
  </bibl>
  
  <bibl id="B50">
    <title><p>Discriminative learning of deep convolutional feature point
    descriptors</p></title>
    <aug>
      <au><snm>Simo Serra</snm><fnm>E</fnm></au>
      <au><snm>Trulls</snm><fnm>E</fnm></au>
      <au><snm>Ferraz</snm><fnm>L</fnm></au>
      <au><snm>Kokkinos</snm><fnm>I</fnm></au>
      <au><snm>Fua</snm><fnm>P</fnm></au>
      <au><snm>Moreno Noguer</snm><fnm>F</fnm></au>
    </aug>
    <source>Proc. ICCV</source>
    <pubdate>2015</pubdate>
    <fpage>118</fpage>
    <lpage>-126</lpage>
  </bibl>
  
  <bibl id="B51">
    <title><p>Learning local feature descriptors using convex
    optimisation</p></title>
    <aug>
      <au><snm>Simonyan</snm><fnm>K</fnm></au>
      <au><snm>Vedaldi</snm><fnm>A</fnm></au>
      <au><snm>Zisserman</snm><fnm>A</fnm></au>
    </aug>
    <source>IEEE PAMI</source>
    <publisher>IEEE</publisher>
    <pubdate>2014</pubdate>
    <volume>36</volume>
    <issue>8</issue>
    <fpage>1573</fpage>
    <lpage>-1585</lpage>
  </bibl>
  
  <bibl id="B52">
    <title><p>Kernel local descriptors with implicit rotation
    matching</p></title>
    <aug>
      <au><snm>Bursuc</snm><fnm>A</fnm></au>
      <au><snm>Tolias</snm><fnm>G</fnm></au>
      <au><snm>J{\'e}gou</snm><fnm>H</fnm></au>
    </aug>
    <source>Proc. ACMM</source>
    <pubdate>2015</pubdate>
    <fpage>595</fpage>
    <lpage>-598</lpage>
  </bibl>
  
  <bibl id="B53">
    <title><p>{NetVLAD}: {CNN} architecture for weakly supervised place
    recognition</p></title>
    <aug>
      <au><snm>Arandjelovic</snm><fnm>R</fnm></au>
      <au><snm>Gronat</snm><fnm>P</fnm></au>
      <au><snm>Torii</snm><fnm>A</fnm></au>
      <au><snm>Pajdla</snm><fnm>T</fnm></au>
      <au><snm>Sivic</snm><fnm>J</fnm></au>
    </aug>
    <source>Proc. CVPR</source>
    <pubdate>2016</pubdate>
    <fpage>5297</fpage>
    <lpage>-5307</lpage>
  </bibl>
  
  <bibl id="B54">
    <title><p>{CNN} Image Retrieval Learns from {BoW}: Unsupervised Fine-Tuning
    with Hard Examples</p></title>
    <aug>
      <au><snm>Radenovi{\'c}</snm><fnm>F</fnm></au>
      <au><snm>Tolias</snm><fnm>G</fnm></au>
      <au><snm>Chum</snm><fnm>O</fnm></au>
    </aug>
    <source>Proc. ECCV</source>
    <pubdate>2016</pubdate>
  </bibl>
  
  <bibl id="B55">
    <title><p>Going deeper with convolutions</p></title>
    <aug>
      <au><snm>Szegedy</snm><fnm>C</fnm></au>
      <au><snm>Liu</snm><fnm>W</fnm></au>
      <au><snm>Jia</snm><fnm>Y</fnm></au>
      <au><snm>Sermanet</snm><fnm>P</fnm></au>
      <au><snm>Reed</snm><fnm>S</fnm></au>
      <au><snm>Anguelov</snm><fnm>D</fnm></au>
      <au><snm>Erhan</snm><fnm>D</fnm></au>
      <au><snm>Vanhoucke</snm><fnm>V</fnm></au>
      <au><snm>Rabinovich</snm><fnm>A</fnm></au>
      <au><cnm>others</cnm></au>
    </aug>
    <pubdate>2015</pubdate>
  </bibl>
  
  <bibl id="B56">
    <title><p>Deep residual learning for image recognition</p></title>
    <aug>
      <au><snm>He</snm><fnm>K</fnm></au>
      <au><snm>Zhang</snm><fnm>X</fnm></au>
      <au><snm>Ren</snm><fnm>S</fnm></au>
      <au><snm>Sun</snm><fnm>J</fnm></au>
    </aug>
    <source>Proc. CVPR</source>
    <pubdate>2016</pubdate>
    <fpage>770</fpage>
    <lpage>-778</lpage>
  </bibl>
  
  <bibl id="B57">
    <title><p>On the analysis and interpretation of inhomogeneous quadratic forms
    as receptive fields</p></title>
    <aug>
      <au><snm>Berkes</snm><fnm>P</fnm></au>
      <au><snm>Wiskott</snm><fnm>L</fnm></au>
    </aug>
    <source>Neural computation</source>
    <publisher>MIT Press</publisher>
    <pubdate>2006</pubdate>
    <volume>18</volume>
    <issue>8</issue>
    <fpage>1868</fpage>
    <lpage>-1895</lpage>
  </bibl>
  
  <bibl id="B58">
    <title><p>Visualizing and understanding convolutional networks</p></title>
    <aug>
      <au><snm>Zeiler</snm><fnm>MD</fnm></au>
      <au><snm>Fergus</snm><fnm>R</fnm></au>
    </aug>
    <source>European conference on computer vision</source>
    <pubdate>2014</pubdate>
    <fpage>818</fpage>
    <lpage>-833</lpage>
  </bibl>
  
  <bibl id="B59">
    <title><p>Object Retrieval with Large Vocabularies and Fast Spatial
    Matching</p></title>
    <aug>
      <au><snm>Philbin</snm><fnm>J.</fnm></au>
      <au><snm>Chum</snm><fnm>O.</fnm></au>
      <au><snm>Isard</snm><fnm>M.</fnm></au>
      <au><snm>Sivic</snm><fnm>J.</fnm></au>
      <au><snm>Zisserman</snm><fnm>A.</fnm></au>
    </aug>
    <source>Proc. CVPR</source>
    <pubdate>2007</pubdate>
  </bibl>
  
  <bibl id="B60">
    <title><p>MatConvNet -- Convolutional Neural Networks for MATLAB</p></title>
    <aug>
      <au><snm>Vedaldi</snm><fnm>A.</fnm></au>
      <au><snm>Lenc</snm><fnm>K.</fnm></au>
    </aug>
    <source>Proc. ACMM</source>
    <pubdate>2015</pubdate>
  </bibl>
  
  <bibl id="B61">
    <title><p>The Yael Library</p></title>
    <aug>
      <au><snm>Douze</snm><fnm>M</fnm></au>
      <au><snm>J{\'e}gou</snm><fnm>H</fnm></au>
    </aug>
    <source>Proc. ACMM</source>
    <publisher>New York, NY, USA: ACM</publisher>
    <series><title><p>MM '14</p></title></series>
    <pubdate>2014</pubdate>
    <fpage>687</fpage>
    <lpage>-690</lpage>
    <url>http://doi.acm.org/10.1145/2647868.2654892</url>
  </bibl>
  
  </refgrp>
  } 

\end{document}